%% file: main.tex
\documentclass[conference,11pt]{IEEEtran}

\usepackage[utf8]{inputenc}
\usepackage[T1]{fontenc}
\usepackage{amsmath,amssymb,amsfonts}
\usepackage{graphicx}
\usepackage{booktabs,siunitx,caption,subcaption,url}
\usepackage{enumitem}
\usepackage{hyperref}
\usepackage{xcolor}
\usepackage{cite}
\usepackage{etoolbox}
\apptocmd{\thebibliography}{\setlength{\itemsep}{0pt}}{}{}

\hypersetup{
  colorlinks=true,
  linkcolor=black,
  citecolor=black,
  urlcolor=black,
}

\setlist{nosep,leftmargin=14pt}
\title{Learnable Total Variation with Lambda Mapping for Low-Dose CT Denoising}

\author{
\IEEEauthorblockN{
Yusuf Talha Basak\textsuperscript{1,2},
Mehmet Ozan Unal\textsuperscript{1,2},
Metin Ertas\textsuperscript{2},
Isa Yildirim\textsuperscript{1,2}
}
\IEEEauthorblockA{
\textsuperscript{1}Istanbul Technical University -- Electronics and Communication Engineering Department, Istanbul, Turkey\\
\textsuperscript{2}Istanbul Technical University -- Biomedical Imaging and AI Lab, Istanbul, Turkey\\
Email: basaky21@itu.edu.tr, unalmehmet@itu.edu.tr, metin.ertas@gmail.com, iyildirim@itu.edu.tr
}
}

\begin{document}
\maketitle

\begin{abstract}
\input{abstract.tex}
\end{abstract}


\section{Introduction}
Reducing radiation exposure in computed tomography (CT) introduces strong quantum noise and streak artifacts, challenging diagnostic accuracy. Traditional Total Variation (TV) regularization~\cite{rudin1992tv} effectively suppresses noise but often oversmooths fine anatomical details. Conversely, deep convolutional neural networks (CNNs)~\cite{zhang2017beyond, jin2017fbpconvnet} yield sharper results but lack physical interpretability and may generate hallucinations.

We propose a Learnable Total Variation (LTV) framework that unifies model-based optimization and deep learning for image-domain LDCT denoising. The method embeds a TV prior into an unrolled primal--dual solver~\cite{chambolle2011firstorder} and employs a U-Net-based LambdaNet to predict a spatially varying regularization map ($\lambda$-map). This coupling enables locally adaptive denoising guided by iterative TV dynamics. The unrolled solver remains fully differentiable, allowing the learned $\lambda$-map to modulate regularization strength: preserving edges while smoothing homogeneous regions.

\vspace{0.5em}
\noindent\textbf{Related Work and Contribution.}
While adaptive weighted TV methods~\cite{liu2012awtv} improve local flexibility by adjusting regularization strength, they still rely on hand-crafted heuristics.
Data-driven approaches like RED-CNN~\cite{chen2017redcnn} and FBPConvNet~\cite{jin2017fbpconvnet} show strong empirical performance but operate as black boxes. Hybrid unrolled methods~\cite{adler2018learned, aggarwal2019modl} address this by embedding learnable components into iterative solvers. Recently, Kofler et al.~\cite{kofler2023ltv} introduced learning spatial regularization maps for reconstruction. Unlike prior work primarily focusing on reconstruction from projections~\cite{kofler2023ltv, morotti2025edgeaware}, our LTV focuses on image-domain denoising using a Chambolle--Pock solver with \emph{learnable} step sizes and relaxation parameters. Furthermore, we demonstrate that for LDCT, minimal distribution-level regularization of the $\lambda$-map yields more stable optimization than enforcing strong structure-aligned priors.

\begin{figure*}[t]
    \centering
    \includegraphics[width=\textwidth]{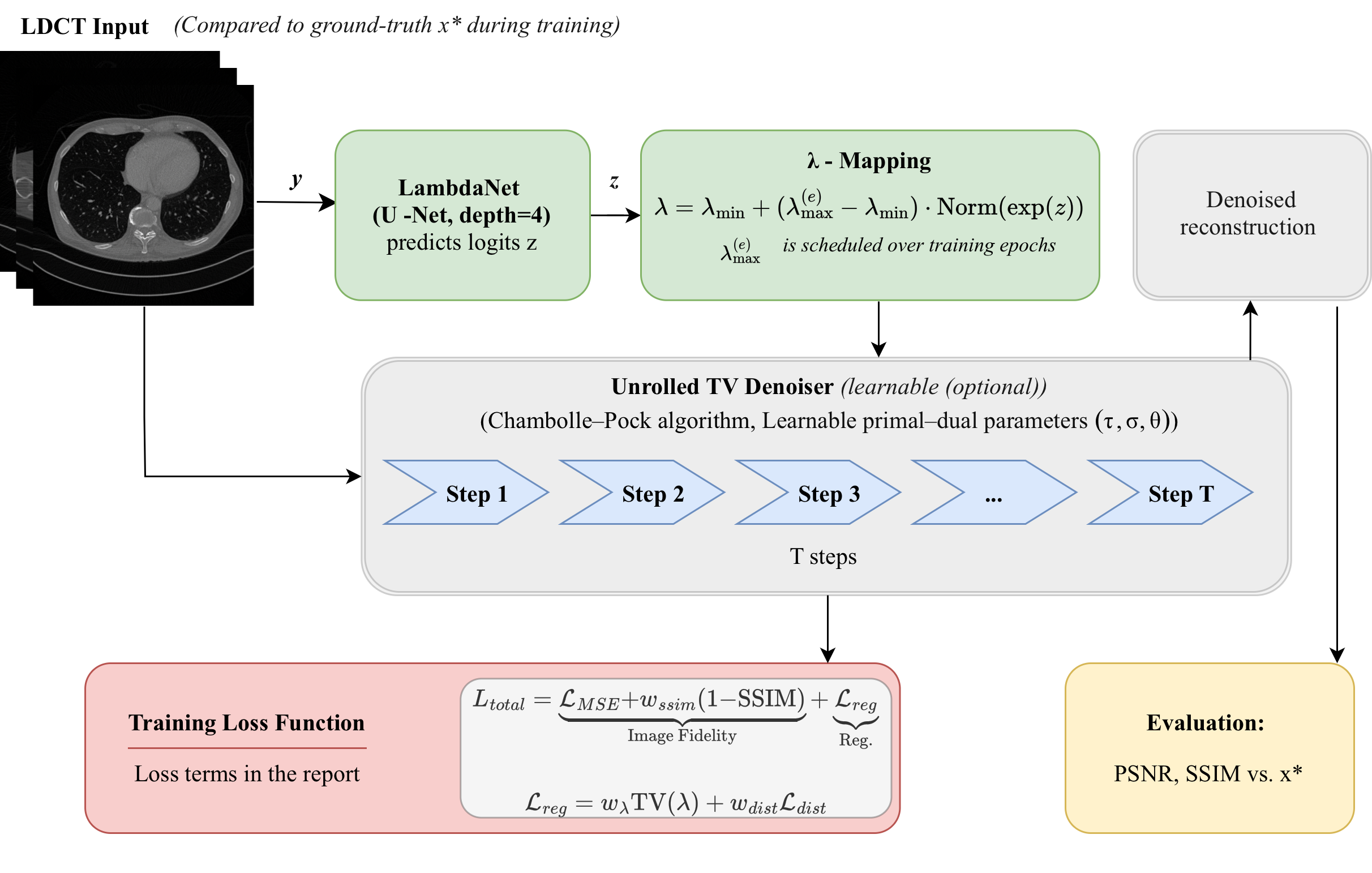}
    \caption{Overview of the proposed LTV framework. A LambdaNet predicts a spatially adaptive $\lambda$-map guiding a $T$-step unrolled Chambolle--Pock solver. The pipeline is trained end-to-end with a composite loss combining image fidelity and $\lambda$-regularization.}
    \label{fig:diagram}
\end{figure*}

\section{Methodology}
\label{sec:method}

\subsection{Overview}
The proposed LTV framework integrates a spatially adaptive regularization network with an unrolled optimization module (Fig.~\ref{fig:diagram}). It consists of (i) \emph{LambdaNet}, predicting a per-pixel $\lambda$-map, and (ii) a differentiable TV denoiser unrolled for $T$ primal--dual updates. The pipeline is trained end-to-end, optimizing reconstruction $\hat{x}$ against ground truth $x^\star$ via gradients propagated through all solver iterations.

\subsection{LambdaNet and Adaptive $\lambda$-Mapping}
\label{ssec:lambdanet}
LambdaNet is a U-Net-style encoder--decoder (four stages, 64 base channels) equipped with attention-gated skip connections. To capture context at varying levels, it employs a multi-scale fusion head that aggregates logits from three decoder stages (full, 1/2, and 1/4 resolution) via a weighted average to produce the logit map $z$.

To ensure a stable dynamic range $[\lambda_{\min}, \lambda_{\max}^{(e)}]$, we adopt an exponential normalization strategy. The pixel-wise regularization map at epoch $e$ is computed as:
\begin{equation}
\label{eq:lambda-map}
\lambda = \lambda_{\min} + \left(\lambda_{\max}^{(e)} - \lambda_{\min}\right) \frac{\tilde{z} - \min(\tilde{z})}{\max(\tilde{z}) - \min(\tilde{z}) + \epsilon},
\end{equation}
where $\tilde{z} = \exp(\mathrm{clip}(z, z_{\min}, z_{\max}))$ and $\epsilon$ is a stability constant. This mapping emphasizes relative differences in logits, preventing trivial collapse while sharpening the contrast between low- and high-regularization regions. Crucially, the upper bound $\lambda_{\max}^{(e)}$ is annealed (e.g., $0.01 \!\rightarrow\! 5.0$) over the first 25 epochs to stabilize early optimization.

\subsection{Unrolled Primal--Dual TV Denoiser}
\label{ssec:solver}
Given the per-pixel map $\lambda$, we use a Chambolle--Pock-style solver unrolled for $T{=}20$ iterations. The solver features learnable step sizes $(\tau,\sigma_d)$ and a relaxation parameter $\theta$, all of which are constrained via softplus and clamping to ensure stable ranges. For ablation and comparison purposes, we also consider a fixed solver configuration in which $(\tau,\sigma_d,\theta)$ are kept constant across all unrolled iterations.
This variant isolates the effect of spatial adaptivity introduced by the learned $\lambda$-map from the additional modeling capacity gained by learning the optimization dynamics themselves.
Unless stated otherwise, the term \emph{LTV (ours)} refers to the fully learnable solver, while \emph{LTV (fixed solver)} denotes this constrained variant. Let $x^{(k)}$ be the primal image and $p^{(k)}$ the dual variable (gradient field).

\textbf{Dual update.} Dual ascent with a pixel-wise projection onto an $\ell_2$ ball defined by $\lambda$:
\begin{equation}
p^{(k+1)} \;=\;
\Pi_{\|\cdot\|\le \lambda}\!\Big(p^{(k)} + \sigma_d \nabla \bar{x}^{(k)}\Big).
\label{eq:dual}
\end{equation}
Here, $\bar{x}^{(k)}$ is the extrapolated primal variable from the previous step. This projection locally bounds the TV penalty based on LambdaNet's prediction.

\textbf{Primal update.} TV-driven smoothing is balanced with data fidelity:
\begin{align}
x^{(k+1)} \;=\;
\frac{x^{(k)} + \tau \nabla^{\!\top} p^{(k+1)} + \tau\,w_{\text{data}}\,y}
{1+\tau\,w_{\text{data}}},
\label{eq:primal}
\end{align}
where $w_{\text{data}} = 1/\sigma_{\text{data}}$ is the data fidelity weight and $\nabla^{\!\top}$ denotes the divergence operator (negative transpose of the gradient $\nabla$).

\textbf{Relaxation.} The iterate is then extrapolated to accelerate convergence:
\begin{equation}
\bar{x}^{(k+1)} \;=\; x^{(k+1)} + \theta\big(x^{(k+1)} - x^{(k)}\big),
\label{eq:relax}
\end{equation}
Here, $\theta$ is a \textbf{learnable over-relaxation} parameter that provides a momentum effect, using the update vector ($x^{(k+1)} - x^{(k)}$) to accelerate convergence. This new iterate $\bar{x}^{(k+1)}$ serves as input to the subsequent dual update. Equations~(\ref{eq:dual})--(\ref{eq:relax}) are explicitly unrolled, making the solver fully differentiable.

\begin{figure*}[t]
    \centering
    \setlength{\tabcolsep}{3pt}
    \renewcommand{\arraystretch}{1.0}

    \begin{tabular}{ccccc}
        \textbf{Ground Truth} &
        \textbf{Noisy} &
        \textbf{TV} &
        \textbf{FBP+U-Net} &
        \textbf{LTV (ours)} \\

        \includegraphics[width=0.19\linewidth]{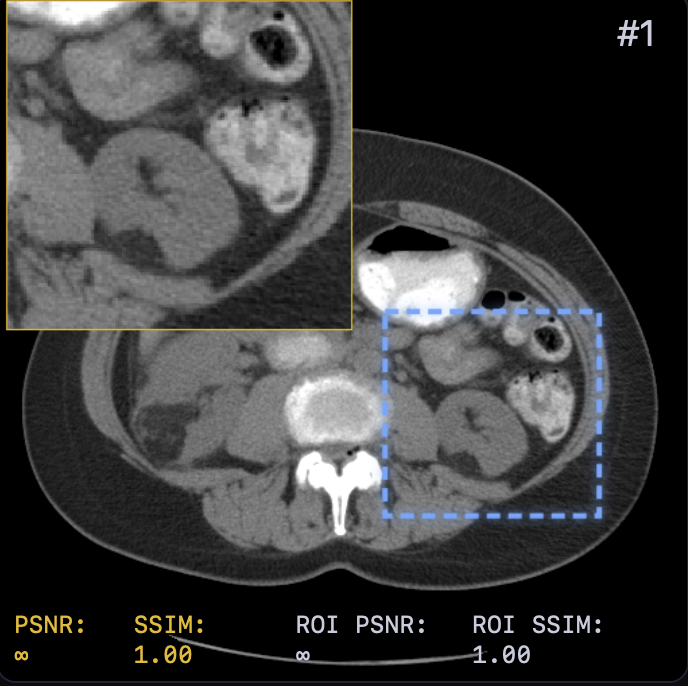} &
        \includegraphics[width=0.19\linewidth]{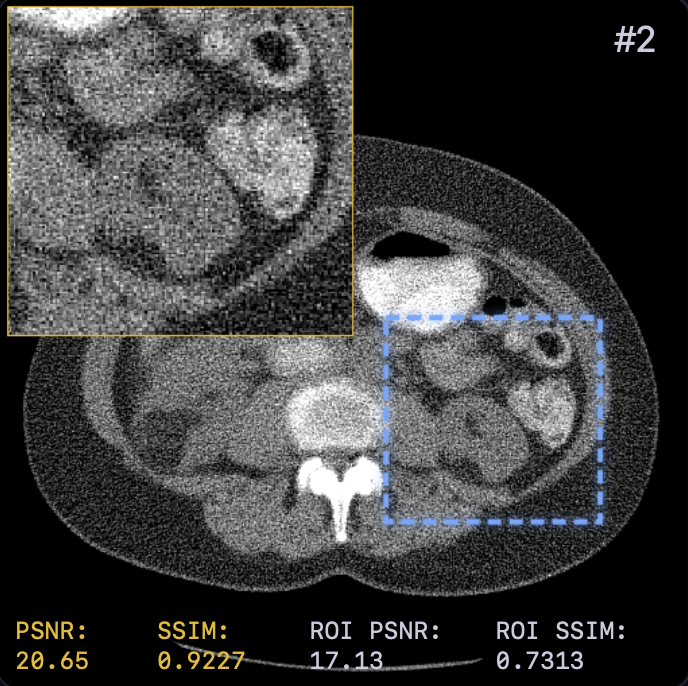} &
        \includegraphics[width=0.19\linewidth]{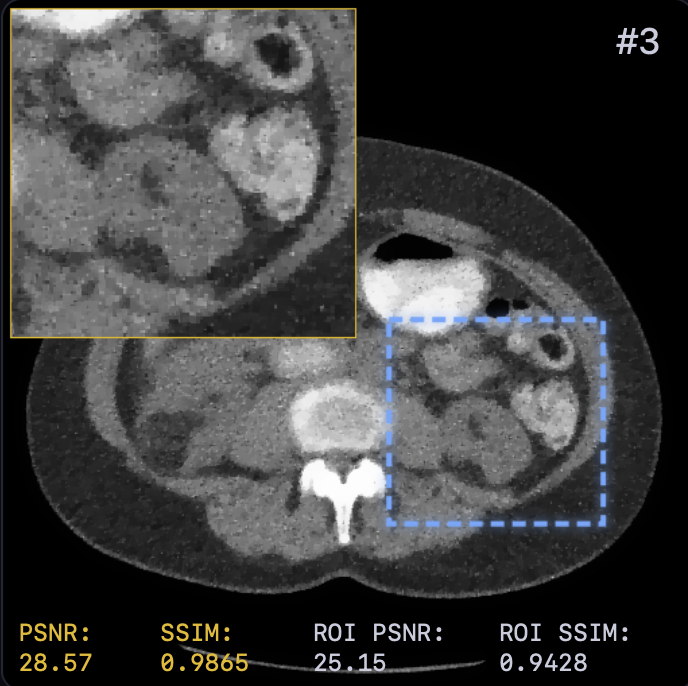} &
        \includegraphics[width=0.19\linewidth]{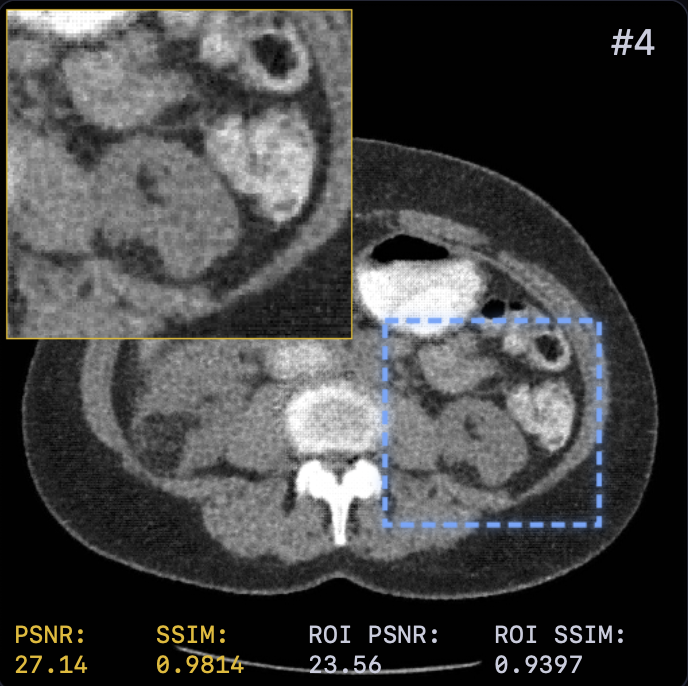} &
        \includegraphics[width=0.19\linewidth]{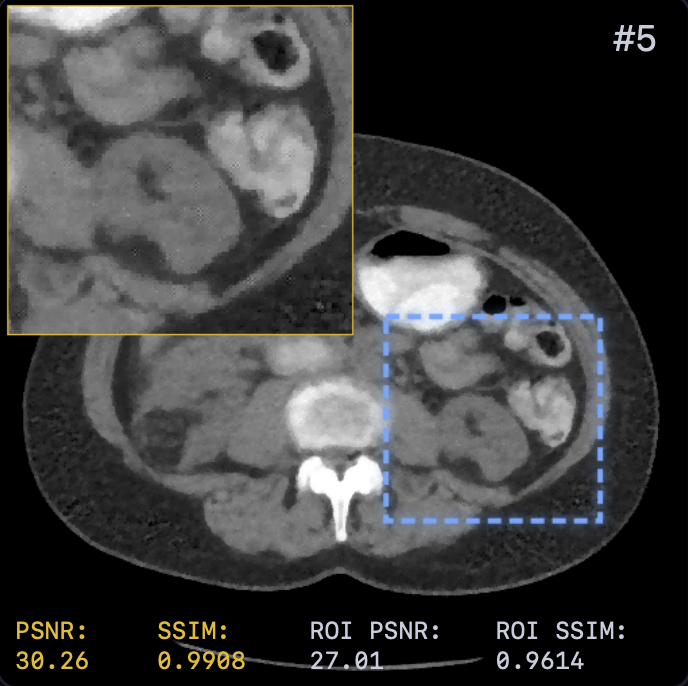} \\

        \includegraphics[width=0.19\linewidth]{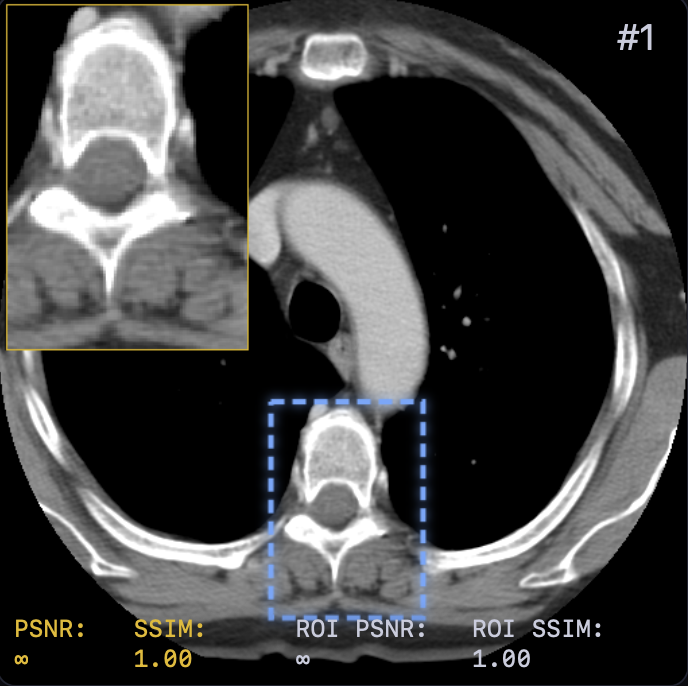} &
        \includegraphics[width=0.19\linewidth]{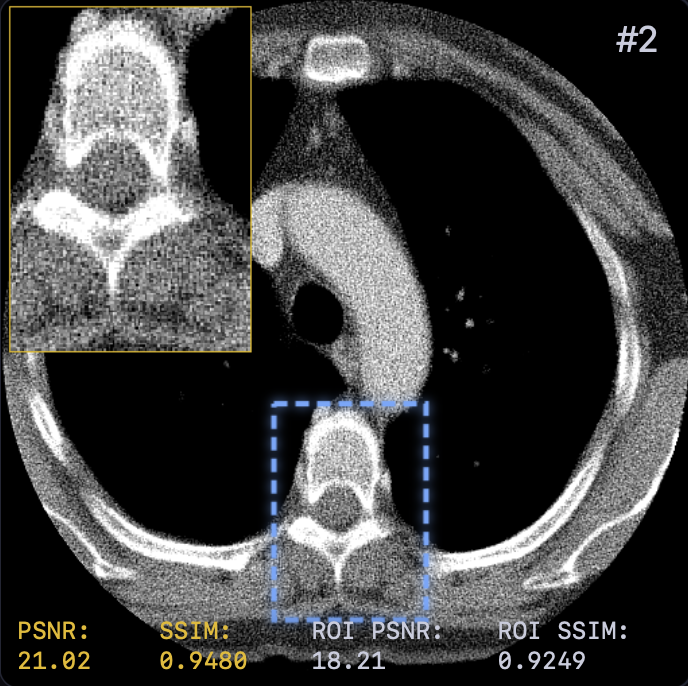} &
        \includegraphics[width=0.19\linewidth]{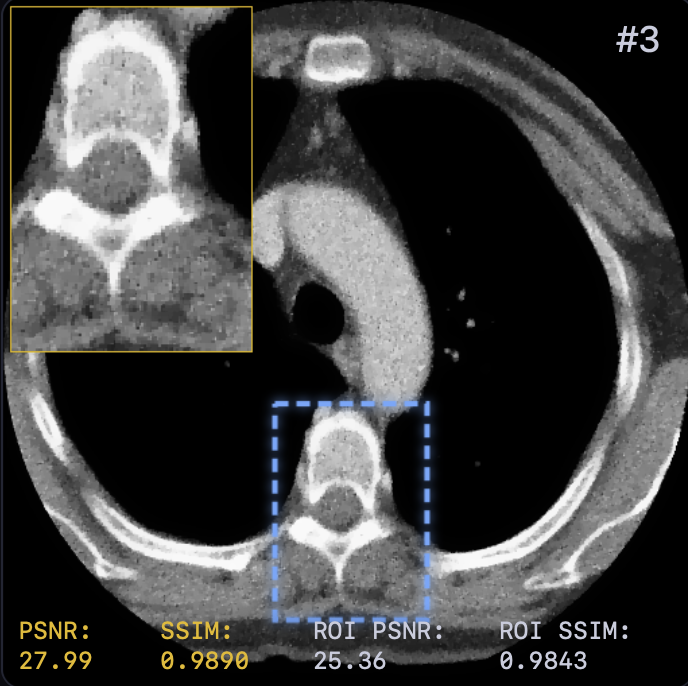} &
        \includegraphics[width=0.19\linewidth]{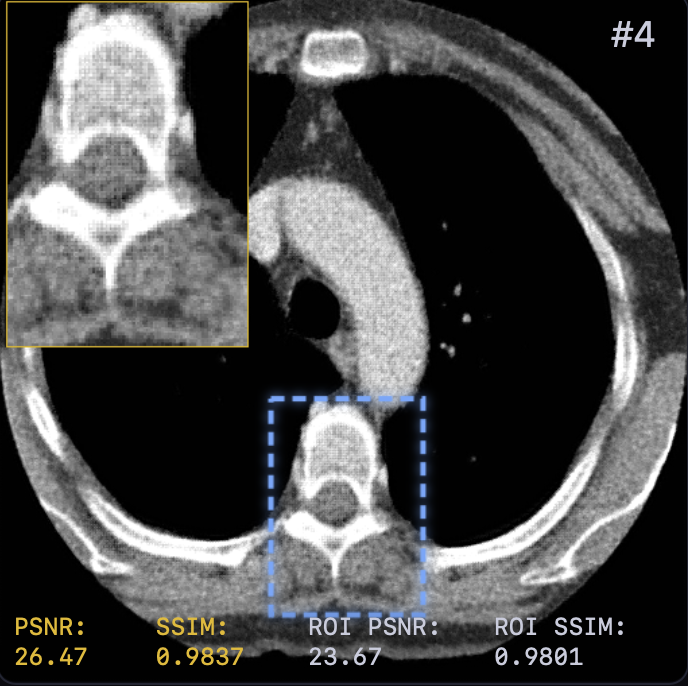} &
        \includegraphics[width=0.19\linewidth]{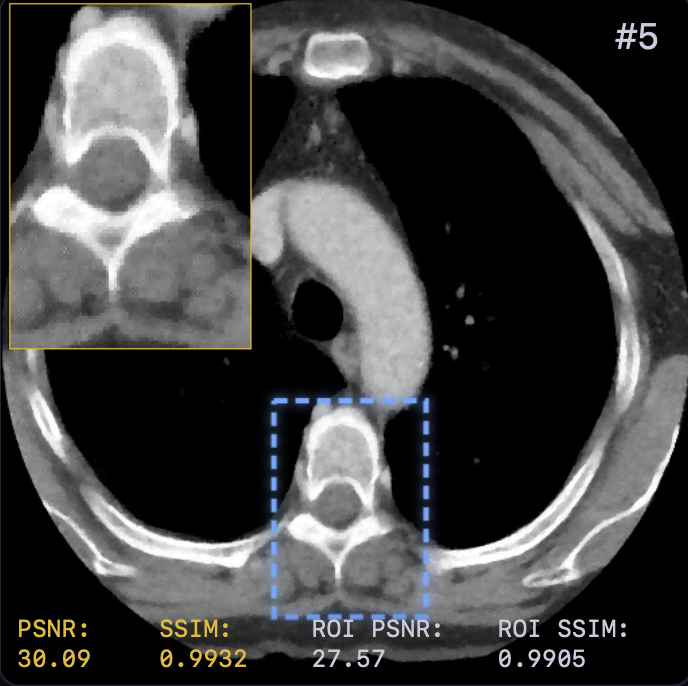} \\
    \end{tabular}

    \caption{Representative LDCT slices comparing noisy, TV, U-Net, and LTV reconstructions. }
    \label{fig:qualitative}
\end{figure*}

\subsection{Training Loss Function}
\label{ssec:loss}

Designing a stable training objective for learning spatially varying regularization maps is non-trivial. While rich structure-aware priors may appear attractive, our empirical analysis revealed that over-constraining the $\lambda$-map often leads to unstable optimization in LDCT denoising. Based on extensive ablation studies, we adopt a \emph{minimal and stability-driven} loss function defined as:

\begin{equation}
\label{eq:loss_final}
\begin{aligned}
\mathcal{L}_{\text{total}}
&= \mathcal{L}_{\text{MSE}}(\hat{x},x^\star)
+ w_{\text{ssim}}\!\left[1-\mathrm{SSIM}(\hat{x},x^\star)\right] \\
&\quad + w_{\lambda}\,\mathrm{TV}(\lambda)
+ w_{\text{dist}}\,\mathcal{L}_{\text{dist}}(\lambda),
\end{aligned}
\end{equation}

The distribution-level regularization term is defined as:

\begin{equation}
\label{eq:ldist}
\mathcal{L}_{\text{dist}}(\lambda) = -\mathrm{Std}(\lambda),
\end{equation}

Here, $\mathrm{Std}(\lambda)$ denotes the standard deviation of the predicted $\lambda$-map. Minimizing the negative standard deviation maximizes the dynamic range, preventing trivial collapse to a constant value.

Image fidelity anchors the reconstruction $\hat{x}$ to the ground truth $x^\star$ using a combination of pixel-wise and structural similarity measures:
\begin{itemize}
    \item $\mathcal{L}_{\text{MSE}}$: Standard $L_2$ loss penalizing pixel-wise intensity errors.
    \item $\mathcal{L}_{\text{SSIM}}$: Structural similarity term capturing contrast and perceptual consistency, which we found crucial for stability compared to pure MSE.
\end{itemize}

To prevent degenerate or spatially noisy $\lambda$-maps while preserving flexibility, we apply two lightweight auxiliary regularizers:
\begin{itemize}
    \item $\mathrm{TV}(\lambda)$: Encourages spatial coherence and suppresses high-frequency artifacts in the $\lambda$-map.
    \item $\mathcal{L}_{\text{dist}}(\lambda)$: A distribution-level regularizer based on variance (\ref{eq:ldist}) to prevent trivial $\lambda$ collapse, promoting a non-trivial dynamic range without enforcing explicit structural alignment.
\end{itemize}

We also explored several structure-aware $\lambda$ priors, including projection-based and edge-aligned constraints. However, ablation results showed that such constraints frequently destabilize optimization and lead to performance degradation. Thus, they are excluded from the final model.

\section{Experiments and Results}
\label{sec:experiments}

\textbf{Datasets and preprocessing.}
We created a paired dataset from DeepLesion~\cite{deeplesion} slices by simulating 10\% dose noise based on the LoDoPaB-CT methodology~\cite{lodopab}. About $1000$ axial slices were split $70/15/15$ by patient. All data were normalized to $[0,1]$ and trained on full $512{\times}512$ axial slices with mild augmentations (rotations, flips, intensity shifts). Low-variance slices were resampled to ensure structural content.

\textbf{Model and training.}
LambdaNet is a 4-level U-Net with attention gates and multi-scale $\lambda$ heads. The $\lambda$-map is constrained to $[\lambda_{\min}, \lambda_{\max}^{(e)}]$, where $\lambda_{\max}$ ramps from $0.01$ to $5$ over the first 25 epochs. Reconstruction uses a 20-step primal–dual TV solver with learnable step-size parameters $(\tau,\sigma_d,\theta)$. Training is end-to-end via Adam (LambdaNet: $5{\times}10^{-4}$; solver: $5{\times}10^{-4}$), batch size $8$, and gradient clipping.\footnote{Our implementation is available at: \url{https://github.com/itu-biai/learnable_tv_for_ldct}}

\subsection{Quantitative and Qualitative Analysis}

Table~\ref{tab:metrics} presents the quantitative comparison. LTV achieves the best overall performance, surpassing FBP+U-Net~\cite{jin2017fbpconvnet} by approximately \textbf{${+}3.7$\,dB} PSNR. Notably, learning the solver dynamics yields a consistent gain ($+0.73$\,dB PSNR) over the fixed-solver baseline (29.42 vs. 30.15\,dB). This confirms that end-to-end optimization of step sizes $(\tau,\sigma_d)$ and relaxation $\theta$ provides modeling capacity beyond spatial adaptivity alone.

\begin{table}[!t]
  \caption{Quantitative comparison on LDCT (mean).}
  \label{tab:metrics}
  \centering
  \begin{tabular}{lcc}
  \toprule
  \textbf{Method} & \textbf{PSNR (dB)} & \textbf{SSIM} \\
  \midrule
  Noisy Input & $23.04$ & $0.704$ \\
  Classical TV & $27.99$ & $0.816$ \\
  FBP+U-Net~\cite{jin2017fbpconvnet} & $26.48$ & $0.784$ \\
  {LTV (fixed solver)} & $29.42$ & $0.841$ \\
  \textbf{LTV (ours)} & $\mathbf{30.15}$ & $\mathbf{0.854}$ \\
  \bottomrule
  \end{tabular}
\end{table}

\begin{figure}[!t]

    \centering
    \includegraphics[width=\columnwidth]{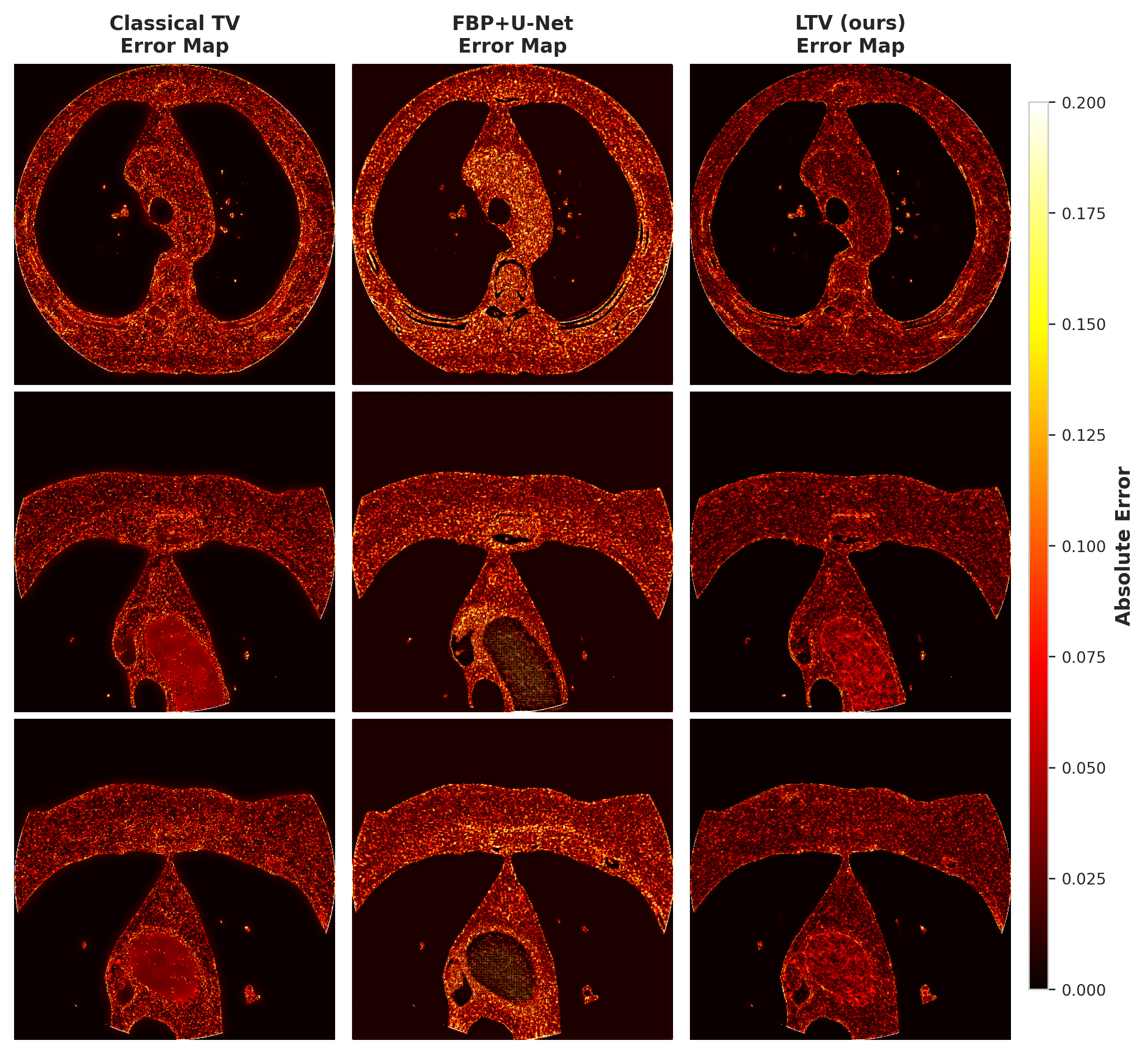}
    \caption{Absolute error maps. LTV yields smaller and more uniform residuals, whereas FBP+U-Net shows high-error zones near edges and lesions.}
    \label{fig:errormap}
\end{figure}

Qualitative results (Fig.~\ref{fig:qualitative}) demonstrate that LTV preserves fine structures (e.g., vessel continuity) while avoiding the over-smoothing of classical TV and the anatomical inconsistencies of FBP+U-Net. This is corroborated by the error maps in Fig.~\ref{fig:errormap}, where LTV exhibits smaller and more spatially uniform residuals, indicating a balanced trade-off between noise suppression and detail preservation.

\subsection{Lambda Adaptivity and Ablation}

The learned $\lambda$-maps exhibit a non-uniform and sparse distribution (mean $\approx 0.18$, median $\sim 10^{-3}$), 
indicating that the model avoids trivial constant regularization. 
We observe a weak but consistent positive correlation with image gradient magnitude ($r \approx 0.15$), 
suggesting that spatial adaptivity emerges naturally from data rather than being explicitly enforced.

Ablation experiments highlight the role of lightweight $\lambda$-map regularization in stabilizing training.
In particular, removing spatial smoothness ($\mathrm{TV}(\lambda)$) or distribution-level regularization leads to noticeable performance degradation 
(approximately \textbf{0.6\,dB} PSNR and \textbf{0.01} SSIM on average), 
and, in some cases, unstable optimization or collapse of the $\lambda$-map.
These results support our design choice of employing minimal, stability-oriented $\lambda$ regularization rather than strong structure-aligned constraints.

\section{Conclusion}
\label{sec:conclusion}

We presented a Learnable Total Variation (LTV) framework that combines an unrolled TV solver with a learned, spatially adaptive $\lambda$-map for low-dose CT denoising. Through extensive ablation studies, we demonstrated that enforcing strong structure-aligned priors on the $\lambda$-map often destabilizes optimization, whereas a minimal, stability-driven training objective leads to robust convergence.

Quantitative and qualitative results on simulated LDCT data show that the proposed approach consistently outperforms classical TV and FBP+U-Net baselines while preserving interpretability through its variational formulation. Crucially, our findings suggest that effective spatial adaptivity emerges naturally from lightweight $\lambda$ regularization combined with end-to-end unrolled optimization, without the need for complex hand-crafted constraints.

This work focuses on image-domain denoising under simulated low-dose conditions, with validation currently limited to a single dataset. Future work will explore extension to volumetric (3D) CT, evaluation on diverse datasets and dose settings, and integration with physics-informed reconstruction pipelines to further assess clinical applicability. Ultimately, the proposed method offers a flexible and effective extension to standard variational models for medical image restoration.

\bibliographystyle{IEEEtran}
\bibliography{refs}
\end{document}

%% file: abstract.tex
While Total Variation (TV) excels in noise reduction and edge preservation, its reliance on a scalar regularization parameter ($\lambda$) limits adaptivity. 
In this study, we present a Learnable Total Variation (LTV) framework coupling an unrolled TV solver with a LambdaNet that predicts a per-pixel regularization map ($\lambda$-map).
The proposed framework is trained end-to-end to optimize reconstruction and regularization jointly, yielding spatially adaptive smoothing. 
Experiments on the DeepLesion dataset, using realistic LoDoPaB-CT simulation, show consistent gains over classical TV and FBP+U-Net, achieving up to \textbf{+3.7\,dB} PSNR and \textbf{8\%} relative SSIM improvement. 
LTV provides an interpretable alternative to black-box CNNs for low-dose CT denoising.